\PassOptionsToPackage{table}{xcolor}
\documentclass{bmvc2k}

\title{Modular Embedding Recomposition for Incremental Learning}

\addauthor{Aniello Panariello}{aniello.panariello@unimore.it}{1}
\addauthor{Emanuele Frascaroli}{emanuele.frascaroli@unimore.it}{1}
\addauthor{Pietro Buzzega}{pietro.buzzega@unimore.it}{1}
\addauthor{Lorenzo Bonicelli}{lorenzo.bonicelli@unimore.it}{1}
\addauthor{Angelo Porrello}{angelo.porrello@unimore.it}{1}
\addauthor{Simone Calderara}{simone.calderara@unimore.it}{1}

\addinstitution{
 AImageLab\\
 University of Modena and Reggio Emilia\\
 Italy
}

\runninghead{Panariello A., \etal}{MoDER}

\usepackage{multirow}
\usepackage{booktabs}
\usepackage{siunitx}
\usepackage{pifont} %
\usepackage{makecell}
\usepackage{algorithm}
\usepackage[noend]{algpseudocode}
\usepackage{cleveref}

\definecolor{lightgray}{gray}{0.95}
\definecolor{midgray}{gray}{0.55}
\definecolor{steelblue}{HTML}{4D82B7}
\definecolor{davysgrey}{rgb}{0.33, 0.33, 0.33}
\definecolor{LightCyan}{rgb}{0.88,1,1}
\definecolor{LightGold}{HTML}{F3E2C5}
\definecolor{AngelRow}{HTML}{FFFDD0}
\definecolor{ao(english)}{rgb}{0.0, 0.5, 0.0}
\definecolor{lightsalmon}{rgb}{1.0, 0.63, 0.48}

\newcommand{\ourcolor}{lightsalmon!20}

\newcommand{\red}[1]{\textcolor{red}{#1}}

\newcommand{\dgreen}[1]{\textcolor{ao(english)}{#1}}

\usepackage[first=0, last=9]{lcg}

\newcommand{\cosinesim}[1]{\ensuremath{\langle #1 \rangle}}

\newcommand{\quotationmarks}[1]{``#1''}

\newcommand{\Star}[1]{#1\ensuremath{^*}\kern-\scriptspace}

\newcommand{\tit}[1]{\smallbreak\noindent\textbf{#1 }}

\newcommand{\moelabel}{\operatorname{MoTE}}
\newcommand{\moelabelv}{\operatorname{-MoTE}}
\newcommand{\vislabel}{\operatorname{vis}}
\newcommand{\textlabel}{\operatorname{txt}}

\newcommand{\vth}{\theta}
\newcommand{\vthptr}{{\theta_0}}

\newcommand{\aphotoof}[1]{\quotationmarks{\texttt{a photo of a [#1]}}}

\newcommand{\textenc}{E^{\textlabel{}}}
\newcommand{\visenc}{E^{\vislabel{}}}

\newcommand{\syndataset}{\mathcal{D_{\operatorname{SYN}}}}
\newcommand{\loss}{\mathcal{L}}

\newcommand{\textemb}{z^{\textlabel{}}}
\newcommand{\visemb}{z^{\vislabel{}}}
\newcommand{\ftemb}{\Tilde{z}^{\textlabel{}}}
\newcommand{\moeemb}{\Tilde{z}^{\moelabel{}}}
\newcommand{\alphamoeemb}{\Tilde{z}^{\alpha\moelabelv{}}}

\newcommand{\seenset}{\mathcal{Y}^{[1:t_c]}}
\newcommand{\topk}{{\operatorname{top}_{\operatorname{K}}}}

\newcommand{\weightmoetemplate}[2]{w_{#1,#2}}
\newcommand{\weightmoe}{\weightmoetemplate{i}{j}}

\newcommand{\taumoe}{{{\pmb{\tau}}_j^{\moelabel{}}}}

\newcommand{\methnam}{{MoDER}\xspace}
\newcommand{\methodname}{{MoDular Embedding Recomposition}\xspace}
\newcommand{\ciltransfer}{{Class Incremental Transfer}\xspace}

\makeatletter
\DeclareRobustCommand\onedot{\futurelet\@let@token\@onedot}
\def\@onedot{\ifx\@let@token.\else.\null\fi\xspace}

\def\eg{\emph{e.g}\onedot} 
\def\ie{\emph{i.e}\onedot}

\def\wrt{w.r.t\onedot} 
\def\etal{\emph{et al}\onedot}
\makeatother

\usepackage{array}
\newcommand{\PreserveBackslash}[1]{\let\temp=\\#1\let\\=\temp}
\newcolumntype{C}[1]{>{\PreserveBackslash\centering}p{#1}}
\newcolumntype{R}[1]{>{\PreserveBackslash\raggedleft}p{#1}}
\newcolumntype{L}[1]{>{\PreserveBackslash\raggedright}p{#1}}

\newcommand{\splitimagenet}{Split Imagenet-R\xspace}
\newcommand{\splitcars}{Split Cars-196\xspace}
\newcommand{\splitcub}{Split CUB-200\xspace}
\newcommand{\spliteurosat}{Split EuroSAT\xspace}

\newcommand{\splitisic}{Split ISIC\xspace}

\newcommand{\shortsplitimagenet}{{\small{ImageNet-R}\xspace}}
\newcommand{\shortsplitcars}{{\small{Cars196}\xspace}}
\newcommand{\shortsplitcub}{{\small{CUB-200}\xspace}}
\newcommand{\shortspliteurosat}{{\small{EuroSAT}\xspace}}

\newcommand{\shortsplitisic}{{\small{ISIC}\xspace}}

\renewcommand{\shortsplitimagenet}{{\small{IN-R}\xspace}}
\renewcommand{\shortsplitcars}{{\small{Cars}\xspace}}
\renewcommand{\shortsplitcub}{{\small{CUB}\xspace}}
\renewcommand{\shortspliteurosat}{{\small{ESAT}\xspace}}

\renewcommand{\shortsplitisic}{{\small{ISIC}\xspace}}

\newcommand{\result}[1]{\ensuremath{#1}}

\newcommand{\faa}[1]{\ensuremath{#1}}

\newcommand{\resultb}[1]{\ensuremath{\mathbf{#1}}}

\newcommand{\numepochsdiff}{30K\xspace}
\newcommand{\optimizer}{AdamW optimizer~\cite{loshchilov2018decoupled}\xspace}
\newcommand{\lrdiff}{\num{1e-3}\xspace}
\newcommand{\wddiff}{\num{1e-2}\xspace}

\newcommand{\embperclass}{15K\xspace}
\newcommand{\batchsizeca}{\num{512}\xspace}

\crefname{section}{Sec.}{Secs.}
\crefname{figure}{Fig.}{Figs.}
\crefname{table}{Tab.}{Tabs.}
\crefname{equation}{Eq.}{Eqs.}
\crefname{algorithm}{Alg.}{Algs.}

\crefname{section}{Section}{Sections}
\crefname{figure}{Figure}{Figures}
\crefname{table}{Table}{Tables}
\crefname{equation}{Equation}{Equations}
\crefname{algorithm}{Algorithm}{Algorithms}

\begin{document}

\maketitle

\begin{abstract}
The advent of pre-trained Vision-Language Models (VLMs) has significantly transformed Continual Learning (CL), mainly due to their zero-shot classification abilities. Such proficiency makes VLMs well-suited for real-world applications, enabling robust performance on novel unseen classes without requiring adaptation. However, fine-tuning remains essential when downstream tasks deviate significantly from the pre-training domain. Prior CL approaches primarily focus on preserving the zero-shot capabilities of VLMs during incremental fine-tuning on a downstream task. We take a step further by devising an approach that transforms preservation into enhancement of the zero-shot capabilities of VLMs. Our approach, named \methodname (\methnam), introduces a modular framework that trains multiple textual experts, each specialized in a single seen class, and stores them in a foundational hub. At inference time, for each unseen class, we query the hub and compose the retrieved experts to synthesize a refined prototype that improves classification. We show the effectiveness of our method across two popular zero-shot incremental protocols, Class-IL and MTIL, comprising a total of 14 datasets. The codebase is available at \href{https://github.com/aimagelab/mammoth}{https://github.com/aimagelab/mammoth}.
\end{abstract}

\section{Introduction}
Vision-Language Models (VLMs) like CLIP~\cite{radford2021learning} have gained significant attention for their flexibility and robust zero-shot capabilities. These characteristics make VLMs particularly appealing in applications demanding high adaptability, where data arrives incrementally, and frequent retraining is not feasible due to time or computational constraints. However, incrementally fine-tuning such models often causes degraded performance in recognizing previously seen classes (\textit{catastrophic forgetting})~\cite{mccloskey1989catastrophic}. Even worse, incrementally fine-tuning VLMs gradually erodes zero-shot capabilities for unseen domains~\cite{zheng2023preventing}. This form of forgetting diminishes the potential to generalize to classes the model has never seen, limiting its adaptability \quotationmarks{out of the box}.

To mitigate this issue, one could resort to Continual Learning (CL), with approaches that generally divide into three categories~\cite{van2019three}. Namely, given a sequence of tasks of disjoint classes, \textit{regularization techniques}~\cite{kirkpatrick2017overcoming,li2017lwf} penalize changes to the most significant weights. \textit{Replay methods}~\cite{rebuffi2017icarl,buzzega2020dark,caccia2022new,mosconi2024mask,millunzi2024may,bellitto2024saliency} interleave past examples, either stored in memory or generated by a model, with new data to reinforce prior knowledge. \textit{Architectural approaches}~\cite{mallya2018packnet,rusu2016progressive} add task-specific parameters to preserve learned representations. While these approaches could be applied to large pre-trained models, the high memory footprint of fine-tuning them makes it impractical. Recent CL methodologies~\cite{smith2023coda,wang2022learning, wang2022dualprompt,frascaroli2024CLIP,wang2023attriclip} incorporate Parameter-Efficient Fine-Tuning (PEFT) techniques~\cite{hu2021lora,zhou2022coop} to reduce such burden. 

In addition to forgetting previously acquired knowledge, fine-tuning a VLM significantly degrades its zero-shot accuracy, as new training data overwrites the broader pre-trained knowledge~\cite{zheng2023preventing,yu2024boosting}. The authors of ZSCL~\cite{zheng2023preventing} introduce this problem under the name of Zero-Shot Continual Learning. However, methods from this field~\cite{zheng2023preventing,yu2024boosting,frascaroli2024CLIP} primarily focus on preventing the deterioration of zero-shot performance and are not designed to actively improve it.

Our approach, \textbf{\methodname} (\textbf{\methnam}), goes further by aiming to \textit{enhance} zero-shot capabilities proactively. To generalize to unseen classes, we train experts incrementally and use them to forge new tailored models on the fly. Specifically, we leverage \textit{model compositionality}~\cite{ortiz-jimenez2023task,mancusi2024is} and devise textual expert modules that are both \textbf{memory-efficient} and \textbf{linearly composable} in parameter space. Thanks to these experts, stored in a repository called the \textbf{foundational hub}, we can forge textual prototypes for unseen classes by reusing models learned across tasks. This provides the basis for building a modular learning system that prevents forgetting and easily adapts to unseen classes and domains.

We validate \methnam on two recognized settings -- Class-Incremental Learning (Class-IL)~\cite{van2019three} and Multi-Domain Task Incremental Learning (MTIL)~\cite{zheng2023preventing,yu2024boosting} -- employing $14$ diverse datasets that present various challenges, including unrelated domains, numerous fine-grained classes, and substantial shifts from the broad pre-training knowledge. To ensure that our proposal improves transfer on unseen classes without sacrificing performance on seen classes, we measure performance as both Class Incremental Transfer~\cite{frascaroli2024CLIP} and Final Average Accuracy. Our main contributions are as follows:
\begin{itemize}
    \item We introduce \methodname, an incremental approach that retains prior knowledge and actively reuses and recombines it to improve zero-shot performance.
    \item We introduce a hub of textual specialized experts, allowing for the creation of new textual prototypes that enhance classification for unseen classes.
    \item We conduct experiments on various datasets and benchmarks showcasing state-of-the-art performance for incremental fine-tuning VLMs.
\end{itemize}

\section{Related Works}
\tit{Class-Incremental Learning and Prompt Tuning.}Learning to Prompt (L2P)~\cite{wang2022learning} pioneers the use of prompts for Class-IL by introducing a shared pool of prompts, with a query function that selects the most relevant prompts for each input image. DualPrompt~\cite{wang2022dualprompt} extends L2P by introducing prefix-tuning~\cite{li2021prefix} and a hierarchy of general task-specific prompts. Similarly, CODA-Prompt~\cite{smith2023coda} applies prefix-tuning and a differentiable, soft mechanism for prompt selection. Among CLIP-based techniques, AttriCLIP~\cite{wang2023attriclip} employs a pool of learnable textual prompts for the CLIP text encoder. Conversely, CGIL~\cite{frascaroli2024CLIP} optimizes a single textual prompt per class and incorporates generated visual features to alleviate forgetting.
\tit{Zero-Shot Continual Learning.}~This framework, recently introduced in~\cite{zheng2023preventing}, addresses the challenge of incrementally fine-tuning Vision-Language Models (VLMs) without compromising their zero-shot performance. To address this, the authors utilize a teacher-student strategy on a reference dataset to preserve pre-training knowledge. Their approach, termed ZSCL, requires substantial GPU memory due to the simultaneous loading of the two CLIP instances. MoE-Adapters~\cite{yu2024boosting} introduces multiple LoRA experts per task to mitigate these limitations. However, this solution necessitates task identity knowledge during inference, adding complexity via a learnable router module that still depends on a reference dataset. CGIL~\cite{frascaroli2024CLIP} employs learnable textual prompts trained on synthetic visual features to prevent forgetting past knowledge; however, its approach for unseen classes is conservative, as it still relies on hand-crafted input templates. Our approach circumvents these constraints by leveraging the composition of class-specific experts from a foundational hub.
\section{Preliminaries}
\tit{Setting.} In Class-Incremental (Class-IL) Continual Learning, a deep model $f(\cdot; \vth)$, parameterized by $\vth$, is trained on a sequence of $T$ tasks. In each one, the model is presented with a training dataset $\mathcal{D}_t = \{(x^{(n)}, y^{(n)})\}_{n=1}^{N_t}$, with $N_t$ examples and labels $y^{(n)}$ drawn from a partition of classes $\mathcal{Y}_t$. In the following, we will refer to the current task with $t_c$ and denote $\seenset = \bigcup_{t=1,\dots,t_c} \mathcal{Y}_t$ as the set of \textbf{seen} classes, namely the union of the classes learned during training, up to and including the current task $t_c$. Importantly, the classes are non-overlapping across tasks, meaning that $\mathcal{Y}_i \cap \mathcal{Y}_j = \emptyset$ for $ i \neq j$. 

The goal in CL is to minimize the cumulative empirical risk across all tasks $\loss_{\text{CL}} = \sum_{t=1}^{T} \ \frac{1}{N_t} \sum_{(x, y) \in \mathcal{D}_t} \left[ \mathcal{L}(f(x; \vth), y) \right]$, where $\loss$ commonly refers to the cross-entropy loss. However, since the model processes tasks sequentially and only has access to the training data of the current task, directly optimizing $\loss_{\text{CL}}$ is impractical. %
\tit{Zero-shot classification with CLIP.} CLIP~\cite{radford2021learning} is composed of two models -- a visual encoder, $\visenc(\cdot)$, and a text encoder, $\textenc(\cdot)$ -- trained to produce aligned representations of paired image-text data. These encoders enable \textit{zero-shot classification}, \ie, categorizing an image $x$ into categories that may not have been seen during training. Specifically, for each candidate class within a set $\mathcal{Y}$, a text prompt is generated by embedding the class name \quotationmarks{\texttt{[CLS]}} within a hand-crafted template such as \aphotoof{CLS}. These class-specific prompts are processed through the text encoder, yielding a textual embedding $\textemb_i$ for the $i$-th class. To determine class logits, cosine similarity $\cosinesim{\cdot,\cdot}$ is computed between the visual embedding $\visemb = \visenc(x)$ of the input image $x$ and each class-level textual embedding.
\tit{Low-Rank Adaption (LoRA).}Fine-tuning methods like LoRA~\cite{hu2021lora} serve to adapt a pre-trained weight matrix $\vthptr^{(l)} \in \mathbf{R}^{d\times k}$ of a layer $l$ in a parameter-efficient manner. In short, given a model with fine-tuned weights $\vth^{\operatorname{FT}}$, LoRA assumes that shifts \wrt pre-training weights $\tau=\vth^{\operatorname{FT}} - \vthptr$ lie on a low-dimensional manifold. Hence, the displacement $\tau$ is parametrized as low-rank matrix $\tau=BA$, where $B\in\mathbf{R}^{d \times r}$, $A\in\mathbf{R}^{r \times k}$, and $r \ll \min(d,k)$. Lowering the rank $r$ can significantly reduce the number of learnable parameters without compromising performance.
\begin{figure*}[t]
    \centering
    \includegraphics[width=0.99\linewidth]{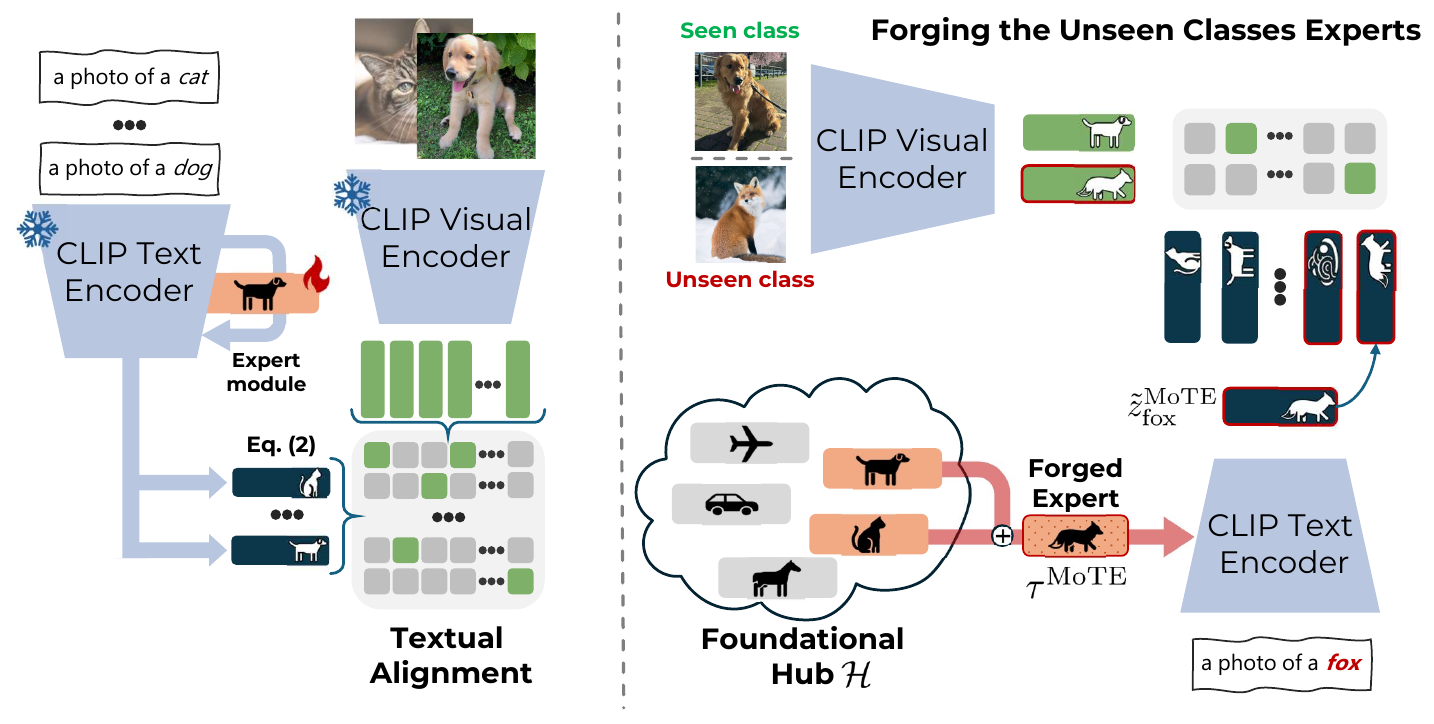}
    \caption{An overview of our approach \methnam. The left side depicts the generative modeling and Textual Alignment (TA) phases. The right side represents the forging of the embedding for the unseen classes.}
    \label{fig:method}
\end{figure*}

\section{\methodname}
The name of our method -- \textbf{\methodname} (\textbf{\methnam}) -- builds on the concept of enhancing capabilities for unseen classes by recomposing fragments of knowledge accumulated across previous tasks. Our training framework consists of specialized experts in the form of lightweight PEFT modules, which fine-tune the CLIP text encoder. We then introduce \textbf{Textual Alignment} (\textbf{TA}, see \cref{sec:atc}), namely a training strategy that promotes the composition of different learnable modules.

During inference, we deploy a dual strategy to compute textual embeddings. Specifically, for classes from the seen set $\seenset$, we use the output of the respective LoRA experts. For the unseen set, we present an approach called \textbf{Mixture of Textual Experts} (\textbf{MoTE}, see~\cref{sec:prototypes}) that leverages compositionality to create experts on the fly. Specifically, given a \textbf{foundational hub} $\mathcal{H}$, comprising all trained experts, we merge the $K$ experts most relevant to the target unseen class.
\subsection{Textual Parameter-Efficient Specialized Experts}
\label{sec:atc}

Given all images $x$ in the current dataset $\mathcal{D}_{t_c}$, we \textbf{freeze} the CLIP visual encoder and extract the relative visual embeddings $\visemb$. These embeddings are used to learn \textbf{aligned} textual prototypes -- one per class in the current task -- ensuring that image-to-text cosine similarity allows for accurate classification. To do this, while existing methods mainly rely on prompt-tuning~\cite{zhou2022coop,zhou2022conditional}, we explore a different approach and devise a distinct LoRA~\cite{hu2021lora} expert module $\tau_i$ for each class. The textual embedding for the class $i$ fine-tuned through the expert is obtained as:
\begin{equation}
\label{eq:embfinetuned}
    \ftemb_i = \textenc(p_i; \vthptr + \tau_i) \quad \text{s.t.} \ \tau_i=B_iA_i \ ,\ i \in \seenset
\end{equation}
where $p_i =$ \aphotoof{CLS} is a hand-crafted textual prompt for the $i$-th class within the seen set. Hence, an expert specializes the CLIP text encoder on input textual prompts related to the same reference class. The displacement $\tau_i$ can be understood as \textbf{task vector}~\cite{ilharco2022editing}, that is, the direction in parameter space along which the capabilities of the pre-trained model rapidly improve for the target class $i$. In a sense, we aim to capture unique aspects of the class $i$ with the corresponding task vector $\tau_i$.

It is noted that, to achieve specialization, each expert in our method builds upon their own set of weights. While this may raise concerns in memory-constrained settings, our formulation is flexible. It can be adapted to define task-specific experts (as shown in~\cref{tab:results_mtil}), requiring only one additional set of parameters per task. Furthermore, we can leverage highly efficient adaptation techniques, such as VeRA~\cite{kopiczko2023vera}. Results for both LoRA- and VeRA-based expert modules are reported in~\cref{tab:main_results}.
\subsection{Textual Alignment (TA)}
While training the textual experts for the newly introduced task, we must consider two aspects: \textit{i)} the new textual prototypes should not interfere with the existing ones, which were trained for previous classes, in order to prevent \textit{catastrophic forgetting} during prediction; \textit{ii)} the experts should be trained to be composable, such that their linear combination yields meaningful outputs. In the following, we will discuss how our training strategy, called Textual Alignment (TA), manages to achieve both of the aforementioned objectives.

To avoid interference between experts and the resulting forgetting issues, we augment the training data for the current task by including highlights from the past, akin to generative replay~\cite{frascaroli2024CLIP}. Namely, we train the experts not only on examples from the current task’s data but also on \textbf{synthetic visual embeddings} generated by lightweight generative models previously trained on past tasks. Specifically, at the onset of the current task, we train a lightweight \textbf{class-conditioned diffusion model}~\cite{ho2020denoising}, denoted by $g_{t_c}(\cdot)$, on the features produced by the visual encoder. Notably, due to the low-dimensional nature of these features, the generator can be implemented as a lightweight Multi-Layer Perceptron (MLP), with a negligible computational cost. In particular, each diffusion model $g_{t_c}(\cdot)$ consists of an eight-layer MLP with $256$ hidden channels per layer and SELU~\cite{klambauer2017selfnormalizing} as the activation function.

Thanks to the generators trained on previous tasks (one per task), we can compensate for the absence of corresponding data in later stages and create balanced training sets for each expert. In practice, after learning the generator for the current task, we construct a synthetic dataset $\syndataset$ by combining the features generated from all the diffusion models. We then train the experts using only data from the $\syndataset$, as discussed in the following.

\tit{Loss function.} To train the PEFT experts, we aim to align the text embeddings in~\cref{eq:embfinetuned} with the visual embeddings in the synthetic dataset $\syndataset$. While a standard cross-entropy loss could be used for multi-modal alignment (as in~\cite{zhou2022coop,radford2021learning}), we treat each class independently by framing the problem as multiple binary classification problems. Specifically, we train the experts with the \textbf{sigmoid loss} $\mathcal{L}_{\sigma}$, similarly to~\cite{zhai2023sigmoid}. Formally, given a data-point $(\visemb, j)$ of class $j$, the per-sample loss becomes:
\begin{align}
&s_i = \cosinesim{\visemb, \ftemb_i}, \ \text{where}\ \ftemb_i \text{is from \cref{eq:embfinetuned}}, \nonumber\\
&\loss_{\sigma} (\visemb, j; \tau_i) = {\sum}_{i \in \seenset} \log(1 + e^{-s_i \cdot \mathbf{1}_{\{i = j\}}}),
\label{eq:loss}
\end{align}
where $\mathbf{1}_{\{i = j\}}$ is an indicator function that returns $1$ if the candidate class $i$ equals the ground truth class $j$ of the synthetic example $\visemb$, and $-1$ otherwise.

We employ the sigmoid loss over cross-entropy primarily to enhance training efficiency. While cross-entropy requires a joint forward/backward pass through all experts, with the sigmoid loss, we can reformulate the problem into multiple independent binary tasks. This allows us to split each update step into manageable batches of experts, which can be distributed across distinct nodes in a multi-GPU or multi-node setup. Finally, as further discussed in the experimental section, we observe that the sigmoid loss yields a consistent yet surprising beneficial effect in terms of transfer to unseen categories.

After their training, the experts are stored in the \textbf{foundational hub} $\mathcal{H}$, which grows incrementally with each task to accommodate new knowledge. Such a hub serves as a deep module library, enabling reuse and composition of expert modules for unseen tasks.
\subsection{Expert Forging via Mixture of Experts}
\label{sec:prototypes}
During evaluation, we compare the visual embeddings extracted from the test images with the textual embeddings from the experts. For a class $i$ of the seen set, we follow \cref{eq:embfinetuned} and prompt the associated expert $\textenc(p_i; \vthptr + \tau_i)$.

Differently, considering a class $j$ from the \textit{unseen set} (\eg, those present in future tasks), we forge a new textual prototype by leveraging the experience accumulated up to the current task and stored in the foundational hub $\mathcal{H}$. Specifically, we first identify the $K$ experts in $\mathcal{H}$ most relevant to the unseen class $j$, then combine their \textbf{weights} to form a new expert. We term this procedure the \textbf{Mixture of Textual Experts} (\textbf{MoTE}); it proceeds as follows::
\begin{equation}
\label{eq:moe}
    \moeemb_{j} = \textenc\left(p_j;\vthptr+\taumoe\right). \text{ where } \taumoe = {\sum}_{i \in \topk(p_j)} \weightmoe \tau_i,
\end{equation}
Here, for an unseen class $j$, $p_j$ denotes its synthetized textual prompt, while $\weightmoe$ represents an affinity score between classes $i$ and $j$. In addition, $\topk(p_j)$ returns the $K$ experts from the seen set $\seenset$ that maximize the similarity \texttt{sim}$(i, j)$. We resort to text-to-text similarity in the original CLIP space, such that \texttt{sim}$(i, j)$ = \cosinesim{\textemb_i, \textemb_j}. We normalize these scores across the $K$ experts with softmax, thus obtaining the $\weightmoe$ in~\cref{eq:moe}.

\tit{Improving expert capabilities.}Following \cref{eq:embfinetuned}, each expert is fed with the same hand-crafted prompt during training. However, this approach will likely result in over-fitting issues, with poor generalization when varying the input prompt. This would be particularly detrimental when forging the textual embedding of an unseen class $j$. Indeed, following~\cref{eq:moe}, the corresponding prompt $p_j$ provided to each expert entails a massive domain shift for the expert $i$, which was trained solely on $p_i$. Therefore, we need a tailored strategy to enhance the robustness of the experts to domain shifts and improve their out-of-distribution capabilities, also shown in~\cite{porrello2024second} to be beneficial for model compositionality.

In this respect, we combine two simple yet effective strategies. The first one is \textbf{\textit{template augmentation}}: in analogy with the concept of data augmentation, we modify the training process by randomly sampling a textual template to construct the input prompt --- \eg, one from the 80 templates commonly used for ImageNet zero-shot tests~\cite{radford2021learning}.

Secondly, inspired by~\cite{wortsman2022robust}, we enhance OOD skills by ensembling the weights $\vth_{i} = \vth + \tau_i$ of each expert with those of the original zero-shot model, represented by $\vthptr$:
\begin{equation}
\label{eq:alpha}
    \vth_{i} = (1-\alpha) \vthptr + \alpha \vth_{i} = \vthptr + \alpha \tau_i,
\end{equation}
where $\alpha \in [0, 1]$. We refer to this step as \textbf{$\alpha$-smoothing} and denote by $\alphamoeemb_j$ the embedding from the \textit{smoothed} experts. With this, the final textual embedding $\alphamoeemb_j$ can be generated with a single forward pass with weights $\vth + \alpha\taumoe$, as follows:
\begin{equation}
\alphamoeemb_j = \textenc(p_j; \vthptr + \alpha \taumoe).
\end{equation}
Training and inference algorithms are provided in \cref{alg:training,alg:evaluation} respectively.
\begin{figure}[t]
  \centering
  \begin{minipage}[t]{0.48\textwidth}
    \centering
    \begin{algorithm}[H]
\caption{\methnam ~ --- ~ Training Phase}\label{alg:training}
\begin{algorithmic}
\smallskip
\For{each dataset $\mathcal{D}_{t_c}, \ {t_c} \in {1,\ldots,T}$}{}
\smallskip
\State Fit $g_{t_c}$ on $ \{ \visenc(x) \mid x \in \mathcal{D}_{t_c} \}$
\State $\syndataset = \{ \visemb \sim g_i , \forall \ i \in {1,\ldots,t_c} \}$
\smallskip
\State \textcolor{blue}{\textit{\textbf{\# Textual Alignment}}}
\For{$\texttt{it} \colon= 1, \dots$}{}
    \State Sample random prompt $p_i \ \forall \ i \in \mathcal{Y}$
    \State $\ftemb_i = \textenc(p_i; \vthptr + \tau_i)$
    \State Sample a batch from $\syndataset$
    \State Update expert $\tau_i$ with $\mathcal{L}_{\sigma}$ (\cref{eq:loss})
\EndFor
\EndFor
\end{algorithmic}
\end{algorithm}
  \end{minipage}
  \hfill
  \begin{minipage}[t]{0.48\textwidth}
    \centering
    \begin{algorithm}[H]
\caption{\methnam ~ --- ~ Forging}\label{alg:evaluation}
\begin{algorithmic}
\For{each class $i$ in $\mathcal{Y}_{\text{seen}}$}{}
\State $\mathcal{Z}[i] \xleftarrow{} \textenc(p_i; \vthptr + \alpha \tau_i)$
\EndFor
\State \textcolor{blue}{\textit{\textbf{\# Mixture of Textual Experts}}}
\For{each class $j$ in $\mathcal{Y}_{\text{unseen}}$}{}
\State $\topk(p_j) \xleftarrow{}$ top-$K$ experts from $\mathcal{H}$
\State $\mathcal{Z}[j]\xleftarrow{}$ Combine their weights
\State $\alphamoeemb_j \xleftarrow{} \textenc(p_j; \vthptr + \alpha \taumoe)$
\EndFor
\State $\visemb \xleftarrow{} \visenc(x_{\text{new}})$ \Comment{\textcolor{blue}{\textit{\textbf{\# classification}}}}
\State $y_{\text{pred}} \xleftarrow{} \arg\max_{y \in \mathcal{Y}} \cosinesim{\visemb, \mathcal{Z}_y}$
\end{algorithmic}
\end{algorithm}
  \end{minipage}
\end{figure}

\section{Experiments}
We evaluate \methnam on Class-Incremental Learning (Class-IL)~\cite{van2019three} and Multi-domain Task Incremental Learning (MTIL)~\cite{zheng2023preventing,yu2024boosting}. Both involve a sequence of image classification tasks, assessing forgetting on the old tasks and generalization to unseen ones. MTIL evaluates transfer capabilities across distinct domains; in Class-IL, the unseen set comes from the same image domain observed during training (\eg, satellite imagery).

\tit{Class-Incremental Learning (Class-IL).} We employ five datasets with varying degrees of alignment with the pre-training. \textit{\splitimagenet}~\cite{hendrycks2021many} (200 classes split into 10 tasks, as in~\cite{wang2022learning,smith2023coda,zhang2023slca}) includes a subset of ImageNet classes, but with a significant domain shift. We also consider two fine-grained datasets, \textit{\splitcars}~\cite{krause20133d} and \textit{\splitcub}~\cite{wah2011cub} (10 tasks each), with only a few samples per class. Additionally, we evaluate on two out-of-distribution datasets: \textit{\spliteurosat}~\cite{helber2018introducing,helber2019eurosat} (RGB satellite images for land cover classification split into 5 binary tasks), and \textit{\splitisic}~\cite{codella2018skin} (6 skin disease classes, split into 3 binary tasks).

\tit{Multi-domain Task Incremental Learning.} MTIL comprises $11$ consecutive tasks, each learning on a distinct dataset: \ie, Aircraft, Caltech101, CIFAR100, DTD, EuroSAT, Flowers, Food, MNIST, OxfordPet, StanfordCars, and SUN397. The datasets are evaluated under two task orders proposed by the original authors (Order-I and Order-II, with our results for Order-II included in the supplementary material). Unlike Class-IL, MTIL relaxes the constraint of unknown task identities at test time. This adjustment simplifies the challenge posed by the \num{1201} classes spread across diverse domains, making evaluation more manageable.

\subsection{Implementation details}

For a fair comparison, all methods use the same backbone. Specifically, in the Class-IL and CI-Transfer setting, we follow~\cite{frascaroli2024CLIP} and employ OpenAI’s CLIP with ViT-L/14 (we also resport results using a ViT-B/16 in the supplementary material). In contrast, in the MTIL setting, we follow~\cite{zheng2023preventing,yu2024boosting} and use OpenAI’s CLIP with ViT-B/16.

Each diffusion model is trained from scratch for \numepochsdiff iterations using the \optimizer with a learning rate of \lrdiff and weight decay of \wddiff. To create the synthetic dataset $\syndataset$, for each class, we sample approximately \embperclass embeddings batched into sizes of \batchsizeca. During the TA phase, we train the LoRA using Adam as optimizer with a learning rate of \num{1e-4} for \splitimagenet, \splitcars and \splitcub, \num{1e-3} for \spliteurosat and \splitisic. The LoRA rank is fixed at $16$. The experiments with VeRA adopt the same hyperparameters, except the learning rates: \num{1e-3} for \splitimagenet, \splitcars and \splitcub, and \num{1e-2} for \spliteurosat and \splitisic.

\subsection{Evaluation metrics}
\begin{table}[t]
\begin{minipage}{.48\linewidth}
  \resizebox{\linewidth}{!}{
  \setlength{\tabcolsep}{2pt}
  
\begin{tabular}{lccccccc}
\midrule
\textbf{Model} & \textbf{\shortsplitimagenet} & \textbf{\shortsplitcars} & \textbf{\shortsplitcub} & \textbf{\shortspliteurosat} & \textbf{\shortsplitisic} & \textbf{Avg.} \\
\midrule
CLIP~\cite{radford2021learning}& \result{82.1} & \result{66.2} & \result{50.9} & \result{55.0} & \result{22.4} & \result{55.3} \\
\midrule
AttriCLIP~\cite{wang2023attriclip}& \result{85.7} & \result{74.0}& \result{54.1} & \result{59.7} & \result{24.1} & \result{61.0} \\
MoE Ad.~\cite{yu2024moe} & \result{88.2} & \result{75.8} & \result{61.7} & \result{55.8} & \result{21.1} & \result{60.5} \\
ZSCL~\cite{zheng2023preventing} & \result{85.3} & \result{72.5} & \result{62.8} & \result{69.7} & \result{25.3} & \result{63.1} \\
CGIL~\cite{frascaroli2024CLIP} & \result{86.7} & \result{78.8} & \result{66.3} & \result{71.5} & \result{48.2} & \result{70.3} \\ \midrule
\textbf{\methnam\scriptsize{{\textcolor{gray}{(LoRA)}}}} & \resultb{89.7} & \resultb{87.0} & \resultb{73.2} & \result{74.0} & \resultb{52.8} & \resultb{75.3} \\
\textbf{\methnam\scriptsize{{\textcolor{gray}{(VeRA)}}}} &\result{89.6} & \result{86.9} & \result{72.9} & \resultb{74.7} & \result{51.7} & \result{75.2} \\
\bottomrule
\end{tabular}%

  }
  \caption{The \ciltransfer on the tested benchmarks.}
  \label{tab:results_transfer}
  \vspace{2em}
  
  \resizebox{\linewidth}{!}{
  \setlength{\tabcolsep}{3pt}
  \begin{tabular}{lcrcrcr}
 \toprule
\textbf{Model} & Transf. & \multicolumn{1}{c}{$\Delta$} & Avg. & \multicolumn{1}{c}{$\Delta$} & Last & \multicolumn{1}{c}{$\Delta$} \\
\midrule
CLIP & \result{69.4} & \textcolor{gray}{$~0.0$} & \result{65.3} & \textcolor{gray}{$~0.0$} & \result{65.3} & \textcolor{gray}{$~0.0$}\\ %
\midrule
Cont. FT & \result{44.6} & \red{-$24.8$} & \result{55.9} & \red{-$9.4$} & \result{77.3} & \dgreen{+$12.0$} \\
LwF & \result{58.9} & \red{-$10.5$} & \result{64.7} & \red{-$0.6$} & \result{74.6} & \dgreen{+$9.3$} \\
Wise-FT & \result{52.3} &\red{-$17.1$} & \result{60.7} & \red{-$4.6$} & \result{77.7} & \dgreen{+$12.4$} \\
ZSCL & \result{68.1} & \red{-$1.3$} & \result{75.4} & \dgreen{+$10.1$} & \result{83.6} & \dgreen{+$18.3$} \\ %
MoE Ad. & \result{68.9} & \red{-$0.5$} & \result{76.7} & \dgreen{+$11.4$} & \result{85.0} & \dgreen{+$19.7$} \\ %
\midrule
\textbf{\methnam} & \resultb{69.7} & \dgreen{\textbf{+}$\boldsymbol{0.3}$} & \resultb{76.9} & \dgreen{\textbf{+}$\boldsymbol{11.6}$}   & \resultb{85.8} & \dgreen{\textbf{+}$\boldsymbol{20.5}$} \\ %
\midrule
\end{tabular}%

  }
  \caption{Accuracy of various approaches in the MTIL setting (Order I).}
  \label{tab:results_mtil}
\end{minipage}
\hfill
\begin{minipage}{.50\linewidth}
\resizebox{\linewidth}{!}{
  \setlength{\tabcolsep}{3pt}
  \begin{tabular}{lcccccc}
\toprule
\textbf{Model} & \textbf{\shortsplitimagenet} & \textbf{\shortsplitcars} & \textbf{\shortsplitcub} & \textbf{\shortspliteurosat} & \textbf{\shortsplitisic} & \textbf{Avg.}\\
\midrule
CLIP~\cite{radford2021learning} & \faa{81.9} & \faa{65.0} & \faa{50.5} & \faa{53.3} & \faa{26.6} & \result{55.5} \\
\midrule
CODA-P.~\cite{smith2023coda} & \result{78.9} & \result{45.2} & \result{72.2} & \result{63.7} & \result{47.4} &\result{61.5} \\
AttriCLIP~\cite{wang2023attriclip} & \result{87.4} & \result{75.6} & \result{58.3} & \result{72.3} & \result{28.3} & \result{64.4} \\
SLCA~\cite{zhang2023slca} & \result{85.5} & \result{73.5} & \resultb{87.8} & \result{93.6} & \result{63.8} & \result{80.8} \\
TMC~\cite{liu2023tangent} & \result{63.2} & \result{39.9} & \result{63.4} & \result{65.0} & \result{48.9} &\result{56.1} \\
Inf-LoRA~\cite{shuo2024inflora} & \result{84.4} & \result{58.0} & \result{80.4} & \result{79.7} & \result{56.1} & \result{71.7}\\
MoE Ad.~\cite{yu2024moe} & \resultb{90.7} & \result{77.8} & \result{65.0} & \result{80.6} & \result{34.5} & \result{69.7} \\
STAR-P.~\cite{menabue2024semantic} & \result{89.2} & \result{86.5} & \result{85.2} & \result{94.2} & \result{67.4} & \result{84.5} \\
ZSCL~\cite{zheng2023preventing} & \result{89.1} & \result{77.7} & \result{62.4} & \result{79.1} & \result{34.1} & \result{68.5} \\
CGIL~\cite{frascaroli2024CLIP} & \result{89.4} & \result{89.3} & \result{83.1} & \result{96.2} & \result{73.0} & \result{86.2} \\\midrule
\textbf{\methnam\scriptsize{{\textcolor{gray}{(LoRA)}}}} & \result{89.7} & \resultb{90.1} & \result{83.7} & \resultb{96.4} & \result{76.3} & \resultb{87.2} \\
\textbf{\methnam\scriptsize{{\textcolor{gray}{(VeRA)}}}} & \result{89.4} & \result{89.6} & \result{82.7} & \result{96.3} & \resultb{76.4} & \result{86.9} \\
\bottomrule
\end{tabular}

  }
  \caption{The Final Avg. Accuracy on the tested benchmarks.}
  \label{tab:main_results}

\end{minipage}
\end{table}

In the Class-Incremental Learning (Class-IL) setting, the task identities remain unknown during inference. To assess performance on unseen classes, we use the \textit{Class Incremantal Transfer} (CI-Transfer) metric~\cite{frascaroli2024CLIP}, which measures accuracy on tasks not yet encountered by the model (see~\cref{tab:results_transfer}). Additionally, we report results on the MTIL benchmark~\cite{zheng2023preventing, yu2024boosting} in~\cref{tab:results_mtil}. For this benchmark, we report the three standard evaluation metrics: \textit{Transfer}, which captures changes in zero-shot accuracy; \textit{Average}, which tracks average accuracy throughout the incremental training; and \textit{Last}, representing accuracy on the final task. Finally, in~\cref{tab:main_results}, we present results in terms of \textit{Final Average Accuracy} scores to assess the impact of the different methods on the seen classes. Results are averaged over three different runs, with standard deviations and further details in the supplementary material.
\subsection{Comparison with the State of the Art}
\label{sec:sota}

\begin{figure*}[t]
    \centering
    \includegraphics[width=0.99\linewidth]{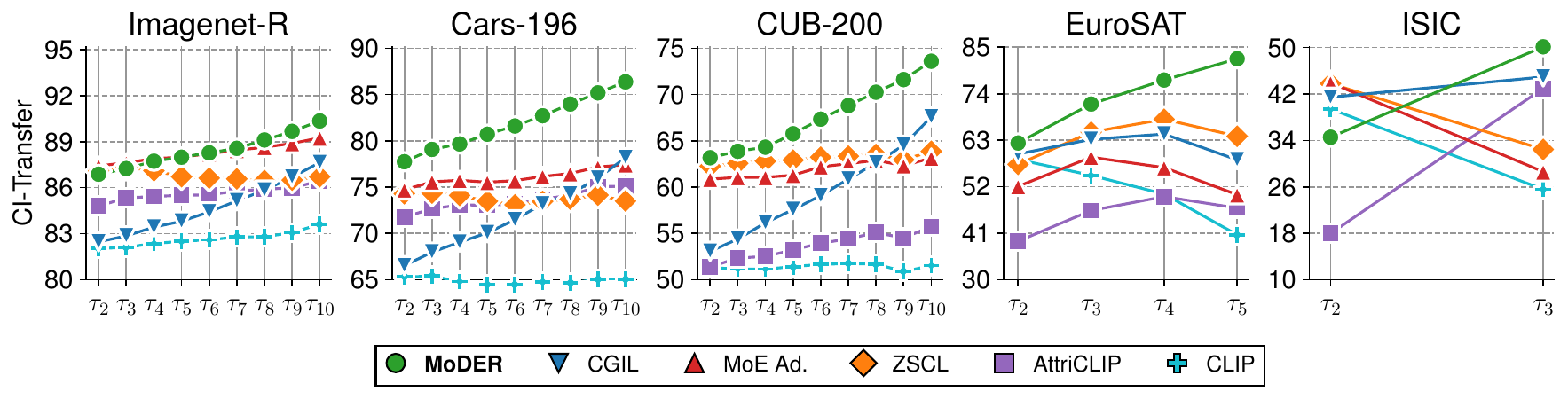}
    \caption{For various benchmarks, the accuracy trend in Class-Incremental transfer indicates the model's effectiveness in transferring to unseen classes in future tasks. A higher trend reflects greater effectiveness in adapting to unseen classes.}
    \label{fig:transfer}
\end{figure*}

\tit{Competing Methods.}
We compare \methnam against several established CL methods from recent literature. We always include a baseline showing the zero-shot performance of CLIP. Following the previous literature, in the MTIL setting we include the Continual FT baseline, which fine-tunes CLIP without any mechanisms for preventing forgetting.
\tit{Results.} Our method, \methnam, demonstrates strong and consistent performance across all benchmarks. In particular, we achieve a clear and substantial lead in terms of CI-Transfer, outperforming all CLIP-based methods (\cref{tab:results_transfer}). Furthermore, \cref{fig:transfer} illustrates the \textit{CI-Transfer} progress throughout training, revealing a steeper increase in zero-shot accuracy compared to other methods, highlighting the effectiveness of our transfer technique.

In the MTIL benchmark, \methnam also excels, surpassing existing approaches across the \textit{Average} and \textit{Last} metrics (\cref{tab:results_mtil}). Notably, it not only prevents zero-shot degradation but improves zero-shot performance, outperforming even CLIP’s original zero-shot accuracy, as shown by the \textit{Transfer} metric. Finally, in the standard Class-IL setting (\cref{tab:main_results}), \methnam achieves the best average performance, albeit by a narrower margin, consistently enhancing both zero-shot generalization to new classes and retention on previously seen ones.

\subsection{Ablative studies}
\label{sec:ablations}

\begin{table}[t]
\noindent \begin{minipage}[t]{.47\linewidth}
  \resizebox{\linewidth}{!}{
  \setlength{\tabcolsep}{2pt}
  
\begin{tabular}{lcccccc}
    \toprule
                               & \textbf{\shortsplitimagenet} & \textbf{\shortsplitcars} & \textbf{\shortsplitcub} & \textbf{\shortspliteurosat} & \textbf{\shortsplitisic} & \textbf{Avg.} \\
    \midrule
    \textbf{\methnam} & \resultb{89.7} & \resultb{87.0} & \resultb{73.2} & \resultb{74.0} & \resultb{52.8} & \resultb{75.3} \\
    \midrule
    \rowcolor{white}
    \rowcolor{lightgray}

    CE loss              & \result{80.9}                & \result{71.2}            & \result{66.8}           & \result{65.7}               & \result{50.8} & \result{67.1}            \\
    \rowcolor{white}
    \midrule
    \rowcolor{white}
    \rowcolor{white}
    No TAug & \result{89.7}     & \result{87.0}                & \result{72.0}            & \result{73.6}           & \result{49.1}               & \result{74.3}            \\
    \midrule
    \rowcolor{lightgray}
    $\alpha=0$                  & \result{88.0}                & \result{86.0}            & \result{72.3}           & \result{55.0}               & \result{34.7}   & \result{67.2}            \\
    \bottomrule
\end{tabular}

  }
  \caption{Ablation analysis on CI-Transfer using LoRA-based experts.}
  \label{tab:ablations_a5}
\end{minipage}
\hfill
\begin{minipage}[t]{.51\linewidth}
  \resizebox{\linewidth}{!}{
  \begin{tabular}{lcc}
\toprule
Model & \textbf{Train. Params} & \textbf{GPU (MiB)} \\
\midrule
LWF~\cite{li2017lwf} & \multicolumn{1}{r}{\num{149.6} M \,($\times$7.1)} & \multicolumn{1}{r}{\num{32172} \,($\times$6.7)} \\
ZSCL~\cite{zheng2023preventing} & \multicolumn{1}{r}{\num{149.6} M \,($\times$7.1)} & \multicolumn{1}{r}{\num{26290} \,($\times$5.5)} \\
MoE Ad.~\cite{yu2024boosting} & \multicolumn{1}{r}{\num{59.8} M \,($\times$2.9)} & \multicolumn{1}{r}{\num{22358} \,($\times$4.7)} \\
\midrule
\textbf{Textual Alignment} & \multicolumn{1}{r}{\num{16.2} M \,($\times$0.8)} & \multicolumn{1}{r}{\num{4748} \,($\times$0.99)} \\
\textcolor{gray}{generator} & \multicolumn{1}{r}{\textcolor{gray}{\num{4.7} M \,($\times$0.2)}} & \multicolumn{1}{r}{\textcolor{gray}{45 \,($\times$0.01)}} \\
\rowcolor{\ourcolor}
\textbf{\methnam} & \multicolumn{1}{r}{\num{20.9} M \,($\times$1)} & \multicolumn{1}{r}{\num{4793} \,($\times$1)} \\
\bottomrule
\end{tabular}

  }
  \caption{Trainable parameters and GPU memory usage for models for MTIL.}
  \label{tab:mtil_computational}
\end{minipage}
\end{table}

We present in~\cref{tab:ablations_a5} the ablation studies on the \ciltransfer metric. 

\tit{Sigmoid loss \textit{vs} cross entropy loss.} Besides the advantage of reducing the computational and memory overhead, we find that the use of the sigmoid loss \textbf{significantly contributes} to the performance of \methnam. This gain is unique to the CI-Transfer metric, as the contribution on the Final Average Accuracy is instead marginal. We conclude that the sigmoid loss, which models each class label independently, provides a more suitable learning objective for experts intended to be composed. Since each class is learned as an independent function, with the sigmoid loss experts can specialize in distinct concepts and be reused modularly in novel combinations. Similarly, Template Augmentation improves performance in CI-Transfer, albeit to a lesser extent than the sigmoid loss.

\tit{Preservation of OOD performance.} The results indicate that $\alpha$-smoothing significantly contribute to the performance on unseen classes. Indeed, when no smoothing is applied ($\alpha=0$), the model achieves inferior performance, especially on out-of-domain datasets.

\tit{On the computational cost of \methnam.} As shown in~\cref{tab:mtil_computational}, \methnam achieves strong performance with substantially fewer trainable parameters and lower GPU memory usage than other baselines. This efficiency highlights its scalability and suitability for large-scale class-incremental scenarios. All results are measured on a single NVIDIA 3060 GPU with 12GB of memory. During \textbf{inference}, \methnam incurs the same overhead as the base CLIP model. Unlike other models such as STAR-Prompt~\cite{menabue2024semantic}, MoE-Adapters~\cite{yu2024moe}, and AttriCLIP~\cite{wang2023attriclip}, which require two forward passes, \methnam requires only a single forward pass on the visual encoder, as the text embeddings can be computed once and cached for future reuse.

\section{Conclusion}
This work builds on the concept of a foundational hub: a virtual, expandable space for incrementally learned models that support the re-use and creation of new expert models. The results in~\cref{sec:sota} demonstrate the effectiveness in mitigating forgetting of previous concepts and facilitating the creation of new ones. Beyond performance, our approach does not incur privacy concerns, is suitable for online scenarios, and is compatible with distributed computing environments since the experts can be trained independently.

\section*{Acknowledgements}
We acknowledge the CINECA award under the ISCRA initiative, for the availability of high performance computing resources and support. This work was supported by the Fortissimo Plus (FFplus) project Grant Agreement No. 101163317), under the European High‑Performance Computing Joint Undertaking (EuroHPC JU) and the Digital Europe Programme. The authors gratefully acknowledge access to compute resources enabled by FFplus. Funded by the European Union.

\bibliography{bib}

\end{document}


\maketitle

\appendix
\pagenumbering{roman}
\renewcommand{\thetable}{\Alph{table}}
\renewcommand{\theequation}{\Alph{equation}}
\renewcommand{\thefigure}{\Alph{figure}}

\section{Extra implementation details}
\label{suppl:implementation}
\subsection{Efficient Batched Training of Experts with Sigmoid loss}
To train the experts, in each task we first collect a synthetic dataset $\syndataset$ and train them jointly (one per class). Theoretically, each update step requires both a forward and a backward pass through all the experts, which can lead to substantial GPU memory demands as the number of classes increases. To address this, we adopt a memory-efficient approach that splits the update step into manageable batches of experts, with the batch size depending on available GPU memory. 

Let $n = 1, \ldots, |\mathcal{Y}_t|$ represent the classes contributing to each backward pass. We begin by computing the text embeddings for the first $n$ classes and then evaluate the objective function in~\cref{eq:loss}. After completing the backward pass for these $n$ classes, we release the GPU memory and compute the loss and backward pass for the following $n$ classes. This cycle continues, with gradients accumulating until all classes have been processed.

\subsection{Further Benchmarks Details}
\label{subsec:benchmarks}
\tit{Backbone.}~To keep fairness, we reproduce all the methods with the same pre-trained CLIP with the ViT-L/14 vision backbone.
\tit{Datasets.}~In the Class-Incremental Learning experiments, we utilized the following datasets:

\begin{itemize}
    \item \textbf{Split Imagenet-R.~\cite{hendrycks2021many}} The dataset is partitioned into 10 tasks containing 20 classes.
    \item \textbf{Split Cars-196.~\cite{krause20133d}} The dataset is divided into 9 tasks of 20 classes each, with the remaining 16 classes assigned to a final task.
    \item \textbf{Split CUB-200.~\cite{wah2011cub}} Similar to Split Imagenet-R, this dataset is divided into 10 tasks of 20 classes each.
    \item \textbf{Split EuroSAT.~\cite{helber2018introducing,helber2019eurosat}} This dataset is organized into 5 tasks, each comprising 2 classes.
    \item \textbf{Split ISIC.~\cite{codella2018skin}} The most frequent class, \emph{Melanocytic nevus}, was removed from the original dataset, making the dataset composed of 6 classes divided into 3 binary tasks.
\end{itemize}

\tit{Image size.} For all benchmarks, the input images (RGB) are rescaled to a resolution of $224 \times 224$.

\tit{Data augmentation.} For methods using CLIP as their backbone, we apply the standard CLIP preprocessing, which includes only RGB normalization. With all other methods, the training phase includes random cropping and horizontal flipping.

\tit{Reproducibility.} Each experiment is conducted three times, using fixed seeds: 1992, 1996, and 1997, as in~\cite{frascaroli2024CLIP}. Each seed defines a unique class order for each dataset, determining how data is partitioned into tasks.

\section{Evaluation Metrics}
\tit{Final Average Accuracy.}~In Class-Incremental Learning, data is presented sequentially across a series of tasks, each with a unique set of class labels that do not overlap. In this setup, the domain typically remains unchanged. The objective is to evaluate the network's ability to learn new tasks without forgetting previously learned ones. To evaluate such performance, we employ the Final Average Accuracy (FAA), the average accuracy across all tasks after completing training. Formally, let $A^i_T$ denote the accuracy on the $i$-th task after training on all $T$ tasks, then FAA is calculated as:
\begin{equation*}
    \operatorname{FAA} = \frac{1}{T} \sum_{i=1}^{T} A^i_T.
\end{equation*}

\tit{CI-Transfer.} To assess performance on unseen tasks, the CI-Transfer~\cite{frascaroli2024CLIP} metric is utilized. This metric represents the cumulative average accuracy on unseen tasks throughout training and is formally defined as:
\begin{equation*}
    \operatorname{CI-Transfer} = \frac{1}{T-1} \sum_{t=1}^{T-1} \left( \frac{1}{T-t} \sum_{i=t+1}^{T} A^i_t \right).
\end{equation*}

\tit{MTIL Metrics.}  
To evaluate our approach, we employ the standard metrics proposed in~\cite{zheng2023preventing}: \textbf{Avg}, \textbf{Last}, and \textbf{Transfer}. We compute such metrics by using the original codebase from~\cite{zheng2023preventing}. As stated in the original paper, the accuracy of these metrics is calculated as in Task Incremental Continual Learning, which means that the model has access to the task-id of the data it is tested on and can only use the respective logits. These metrics measure the model's ability to retain knowledge from pre-training and downstream tasks.

\textbf{Avg} quantifies the average accuracy across all datasets at all time steps:
\begin{equation*}
    \operatorname{Avg} = \frac{1}{T} \sum_{i=1}^{T} A^i, \quad \text{where} \ \ A^i = \frac{1}{T} \sum_{t=1}^{T} A_t^i,
\end{equation*}
where $A_t^i$ is the accuracy of the $i$-th domain after training on task $t$, and $T$ is the total number of tasks.

\textbf{Last} captures the average accuracy across all datasets at the final time step. The formula is the same as the Final Average Accuracy computed in the Class-Incremental Learning setting. However, the single accuracies are evaluated in the Task Incremental scenario.

\textbf{Transfer} evaluates zero-shot performance on unseen domains:
\begin{equation*}
    \operatorname{Transfer} = \frac{1}{T-1} \sum_{i=2}^{T} \text{TR}^i, \ \text{where} \ \ \text{TR}^i = \frac{1}{i-1} \sum_{t=1}^{i} A_t^i.
\end{equation*}
\section{Standard Deviations}
\label{suppl:std}
\cref{tab:main_stds,tab:transfer_stds} present the standard deviation of different runs of our experiments. As stated in \cref{subsec:benchmarks}, each run of the same dataset has a different order of classes (which is consistent across all methods). Consequently, the performance can vary significantly, especially in datasets with a lower number of classes (\textit{Split EuroSAT} and \textit{Split ISIC}). With the \ciltransfer metric, this variance is further exacerbated by how the metric is calculated, with later classes having more weight on the final value. This behaviour also affects the performance of CLIP zero-shot, despite the model being frozen.

\begin{table*}[t]
\rowcolors{2}{lightgray}{}
\begin{center}
\begin{tabular}{lccccc}
\toprule
\textbf{Model} & \textbf{\shortsplitimagenet} & \textbf{\shortsplitcars} & \textbf{\shortsplitcub} & \textbf{\shortspliteurosat} & \textbf{\shortsplitisic} \\
\midrule
CODA-P.~\cite{smith2023coda} & \resultSTD{0.30} & \resultSTD{2.29} & \resultSTD{1.03} & \resultSTD{0.95} & \resultSTD{1.83} \\
AttriCLIP~\cite{wang2023attriclip} & \resultSTD{0.41} & \resultSTD{0.06} & \resultSTD{1.21} & \resultSTD{2.09} & \resultSTD{1.07} \\
SLCA~\cite{zhang2023slca} & \resultSTD{0.12} & \resultSTD{0.19} & \resultSTD{0.06} & \resultSTD{0.78} & \resultSTD{5.02} \\
TMC~\cite{liu2023tangent} & \resultSTD{0.38} & \resultSTD{1.24} & \resultSTD{1.70} & \resultSTD{5.74} & \resultSTD{1.76} \\
Inf-LoRA~\cite{shuo2024inflora} & \resultSTD{0.13} & \resultSTD{1.04} & \resultSTD{0.53} & \resultSTD{3.81} & \resultSTD{2.04} \\
MoE Ad.~\cite{yu2024moe} & \resultSTD{0.15} & \resultSTD{1.02} & \resultSTD{0.29} & \resultSTD{0.53} & \resultSTD{8.25} \\
STAR-P.~\cite{menabue2024semantic} & \resultSTD{0.04} & \resultSTD{0.20} & \resultSTD{0.28} & \resultSTD{0.15} & \resultSTD{0.62} \\
ZSCL~\cite{zheng2023preventing} & \resultSTD{0.41} & \resultSTD{0.18} & \resultSTD{0.11} & \resultSTD{0.48} & \resultSTD{5.34} \\
CGIL~\cite{frascaroli2024CLIP} & \resultSTD{0.12} & \resultSTD{0.14} & \resultSTD{0.10} & \resultSTD{0.10} & \resultSTD{1.75} \\
\midrule
\rowcolor{\ourcolor}
\textbf{\methnam\scriptsize{{\textcolor{gray}{(LoRA)}}}} & \resultSTD{0.2} & \resultSTD{0.3} & \resultSTD{0.3} & \resultSTD{0.1} & \resultSTD{1.2} \\
\textbf{\methnam\scriptsize{{\textcolor{gray}{(VeRA)}}}} & \resultSTD{0.3} & \resultSTD{0.2} & \resultSTD{0.2} & \resultSTD{0.2} & \resultSTD{0.8} \\
\bottomrule
\end{tabular}
\end{center}
  \caption{Standard deviations on Class-IL benchmark (\cref{tab:main_results}).}
  \label{tab:main_stds}
\end{table*}

\begin{table}[t]
\rowcolors{2}{lightgray}{}
\begin{center}
\begin{tabular}{lccccc}
\toprule
\textbf{Model} & \textbf{\shortsplitimagenet} & \textbf{\shortsplitcars} & \textbf{\shortsplitcub} & \textbf{\shortspliteurosat} & \textbf{\shortsplitisic} \\
\midrule
CLIP~\cite{radford2021learning}& \resultSTD{1.49} & \resultSTD{1.18} & \resultSTD{1.69} & \resultSTD{14.08} & \resultSTD{10.92} \\
\midrule
AttriCLIP~\cite{wang2023attriclip}& \resultSTD{1.14} & \resultSTD{0.89}& \resultSTD{1.73} & \resultSTD{9.10} & \resultSTD{13.37} \\
MoE Ad.~\cite{yu2024moe} & \resultSTD{1.09} & \resultSTD{1.20} & \resultSTD{1.04} & \resultSTD{4.08} & \resultSTD{12.21} \\
ZSCL~\cite{zheng2023preventing} & \resultSTD{1.79} & \resultSTD{0.71} & \resultSTD{1.15} & \resultSTD{6.49} & \resultSTD{11.56} \\
CGIL~\cite{frascaroli2024CLIP} & \resultSTD{0.77} & \resultSTD{0.79} & \resultSTD{1.10} & \resultSTD{9.83} & \resultSTD{5.42} \\ \midrule
\rowcolor{\ourcolor}
\textbf{\methnam\scriptsize{{\textcolor{gray}{(LoRA)}}}} & \resultSTD{1.5} & \resultSTD{0.5} & \resultSTD{1.2} & \resultSTD{10.8} & \resultSTD{4.2} \\
\textbf{\methnam\scriptsize{{\textcolor{gray}{(VeRA)}}}} & \resultSTD{1.5} & \resultSTD{0.5} & \resultSTD{1.2} & \resultSTD{9.5} & \resultSTD{3.3} \\
\bottomrule
\end{tabular}
\end{center}
    \caption{Standard deviations on CI-Transfer benchmark (\cref{tab:results_transfer}).}
    \label{tab:transfer_stds}
\end{table}

\section{Additional results}
\begin{table*}
    \centering
    \resizebox{\linewidth}{!}{
    \begin{tabular}{cL{3cm}C{1cm}C{1cm}C{1cm}C{1cm}C{1cm}C{1cm}C{1cm}C{1cm}C{1cm}C{1cm}C{1cm}C{1.5cm}}
        \toprule
           &  Method & \makecell[c]{\rotatebox{90}{Aircraft}} & \makecell[c]{\rotatebox{90}{Caltech101}} & \makecell[c]{\rotatebox{90}{CIFAR100}} & \makecell[c]{\rotatebox{90}{DTD}} & \makecell[c]{\rotatebox{90}{EuroSAT}} & \makecell[c]{\rotatebox{90}{Flowers}} & \makecell[c]{\rotatebox{90}{Food}} & \makecell[c]{\rotatebox{90}{MNIST}} & \makecell[c]{\rotatebox{90}{OxfordPet}} & \makecell[c]{\rotatebox{90}{Cars}} & \makecell[c]{\rotatebox{90}{SUN397}} & \makecell[c]{{\textit{Avg.}}} \\
  
        \midrule
        
            \multirow{3}{*}{\rotatebox{90}{CLIP}}& Zero-shot & 24.3 & 88.4 & 68.2 & 44.6 & 54.9 & 71.0 & 88.5 & 59.4 & 89.0 & 64.7 & 65.2 & 65.3  \\
            & Full Fine-tune & 62.0 & 95.1 & 89.6 & 79.5 & 98.9 & 97.5 & 92.7 & 99.6 & 94.7 & 89.6 & 81.8 & 89.2  \\
            & Fine-tune Adapter & 56.8 & 92.6 & 89.4 & 79.0 & 98.4 & 97.0 & 92.9 & 99.2 & 94.1 & 89.1 & 82.7 & 88.3 \\
            
            \midrule\midrule
            \multirow{6}{*}{\rotatebox{90}{\textbf{Transfer}}} &Continual-FT & & 67.1 & 46.0 & 32.1 & 35.6 & 35.0 & 57.7 & 44.1 & 60.8 & 20.5 & 46.6 & 44.6 \\
            & LwF \cite{li2017lwf} &  & 74.5 & 56.9 & 39.1 &\underline{51.1} & 52.6 & 72.8 & \underline{60.6} & 75.1 & 30.3 & 55.9 & 58.9 \\
            & WiSE-FT \cite{wortsman2022robust} & & 73.5 & 55.6 & 35.6 & 41.5 & 47.0 & 68.3 & 53.9 & 69.3 & 26.8 & 51.9 & 52.3  \\
            & ZSCL \cite{zheng2023preventing} & & 86.0 & 67.4 &\textbf{45.4} & 50.4 & 69.1 & 87.6 &\textbf{61.8} & \underline{86.8} & 60.1 &\textbf{66.8} & 68.1 \\
            & MoE Ad. \cite{yu2024boosting} &&\underline{87.9} & \underline{68.2} & 44.4 &49.9 &\underline{70.7} & \underline{88.7} &59.7 &\textbf{89.1} &\underline{64.5} &65.5 &\underline{68.9}\\
            \rowcolor{\ourcolor}
            & \textbf{\methnam} & & \textbf{88.6} & \textbf{68.5} & \underline{45.2} & \textbf{56.6} & \textbf{71.5} & \textbf{88.9} & 58.8 & \textbf{89.1} & \textbf{64.8} & \underline{65.6}  & \textbf{69.7}\\
            \midrule 
            \multirow{6}{*}{\rotatebox{90}{\textbf{Average}}} &Continual-FT    & 25.5 & 81.5 & 59.1 & 53.2 & 64.7 & 51.8 & 63.2 & 64.3 & 69.7 & 31.8 & 49.7 & 55.9 \\
            & LwF \cite{li2017lwf} & 36.3 & 86.9 & 72.0 & 59.0 & 73.7 & 60.0 & 73.6 & \underline{74.8} & 80.0 & 37.3 & 58.1 & 64.7 \\
            & WiSE-FT \cite{wortsman2022robust} & 26.7 & 86.5 & 64.3 & 57.1 & 65.7 & 58.7 & 71.1 & 70.5 & 75.8 & 36.9 & 54.6 & 60.7  \\
            & ZSCL \cite{zheng2023preventing} & 45.1 & \underline{92.0} & \underline{80.1} & 64.3 & \underline{79.5} & 81.6 & \underline{89.6} & \textbf{75.2} & 88.9 & 64.7 & \textbf{68.0} &75.4 \\
            & MoE Ad. \cite{yu2024boosting} &\underline{50.2}&91.9&\textbf{83.1}&\textbf{69.4}&78.9&\underline{84.0}&89.1&73.7&\underline{89.3}&\underline{67.7}&66.9&\underline{76.7}\\
            \rowcolor{\ourcolor}
            & \textbf{\methnam} & \textbf{51.5} & \textbf{96.0} & 77.7 & \underline{67.4} & \textbf{79.9} & \textbf{85.2} & \textbf{90.0} & 71.9 & \textbf{90.4} & \textbf{68.6} & \underline{67.0}  & \textbf{76.9}\\
            \midrule
            \multirow{6}{*}{\rotatebox{90}{\textbf{Last}}} &Continual-FT & 31.0 & 89.3 & 65.8 & 67.3 & 88.9 & 71.1 & 85.6 & \textbf{99.6} & 92.9 & 77.3 & \underline{81.1} & 77.3 \\
            & LwF \cite{li2017lwf} & 26.3 & 87.5 & 71.9 & 66.6 & 79.9 & 66.9 & 83.8 & \textbf{99.6} & 92.1 & 66.1 & 80.4 & 74.6 \\
            & WiSE-FT \cite{wortsman2022robust} & 27.2 & 90.8 & 68.0 & 68.9 & 86.9 & 74.0 & 87.6 & \textbf{99.6} & 92.6 & 77.8 & \textbf{81.3} & 77.7  \\
            & ZSCL \cite{zheng2023preventing} & 40.6 & \underline{92.2} & \underline{81.3} & 70.5 & \underline{94.8} & 90.5 & \textbf{91.9} & \underline{98.7} & \textbf{93.9} & \underline{85.3} & 80.2 & 83.6 \\
            & MoE Ad. \cite{yu2024boosting} & \underline{49.8}&\underline{92.2}&\textbf{86.1}&\textbf{78.1}&\textbf{95.7}&\underline{94.3}&89.5&98.1&89.9&81.6&80.0&\underline{85.0}\\
            \rowcolor{\ourcolor}
            & \textbf{\methnam} & \textbf{52.3} & \textbf{96.7} & 80.4 & \underline{75.7} & 93.6 & \textbf{96.7} & \underline{91.1} & 95.7 & \underline{93.8} & \textbf{86.2} & \underline{81.1} & \textbf{85.8}\\
        \bottomrule
    \end{tabular}}
    \caption{Comparison with state-of-the-art methods on the MTIL benchmark with Order I. All methods use CLIP ViT-B/16. We highlight the best and second best with \textbf{bold} and \underline{underline} styles respectively.}
    \label{tab:mtil_full_1}
\end{table*}

\begin{table*}
    \centering
    \resizebox{\linewidth}{!}{
    \begin{tabular}{cL{3cm}C{1cm}C{1cm}C{1cm}C{1cm}C{1cm}C{1cm}C{1cm}C{1cm}C{1cm}C{1cm}C{1cm}C{1.5cm}}
        \toprule
                      & {\hspace{1em}} Method & \makecell[c]{\rotatebox{90}{Cars}} & \makecell[c]{\rotatebox{90}{Food}} & \makecell[c]{\rotatebox{90}{MNIST}} & \makecell[c]{\rotatebox{90}{OxfordPet}} & \makecell[c]{\rotatebox{90}{Flowers}} & \makecell[c]{\rotatebox{90}{SUN397}} & \makecell[c]{\rotatebox{90}{Aircraft}} & \makecell[c]{\rotatebox{90}{Caltech101}} & \makecell[c]{\rotatebox{90}{DTD}} & \makecell[c]{\rotatebox{90}{EuroSAT}} & \makecell[c]{\rotatebox{90}{CIFAR100~}} & \makecell[c]{{\textit{Avg.}}} \\
  
        \midrule
        
            \multirow{3}{*}{\rotatebox{90}{CLIP}}& Zero-shot & 64.7 & 88.5 & 59.4 & 89.0 & 71.0 & 65.2 & 24.3 & 88.4 & 44.6 & 54.9 & 68.2 & 65.3  \\
            & Full Fine-tune & 89.6 & 92.7 & 99.6 & 94.7 & 97.5 & 81.8 & 62.0 & 95.1 & 79.5 & 98.9 & 89.6 & 89.2  \\
            & Fine-tune Adapter & 89.1 & 92.9 & 99.2 &94.1&97.0&82.7&56.8&92.6&79.0&98.4&89.4&88.3\\
            
            \midrule\midrule
            \multirow{6}{*}{\rotatebox{90}{\textbf{Transfer}}} &Continual-FT & & 85.9&\underline{59.6}&57.9&40.0&46.7&11.1&70.0&30.5&26.6&37.7& 46.6 \\
            & LwF \cite{li2017lwf} &&87.8&58.5&71.9&46.6&57.3&12.8&81.4&34.5&34.5&46.8& 53.2  \\
            & WiSE-FT \cite{wortsman2022robust} & &87.2 & 57.6 & 67.0 & 45.0 & 54.0 & 12.9 & 78.6 & 35.5 & 28.4 & 44.3 &   51.1\\
            & ZSCL \cite{zheng2023preventing} & & 88.3 & 57.5 & 84.7 & 68.1 & \underline{64.8} & \underline{21.1} & \underline{88.2} & \textbf{45.3} & \underline{55.2} & \textbf{68.2} & 64.1  \\
            & MoE Ad. \cite{yu2024boosting} & & \textbf{88.8} & 59.5 & \underline{89.1} & \underline{69.9} & 64.4 & 18.1 & 86.9 & 43.7 & 54.6 & \textbf{68.2} & \underline{64.3} \\
            \rowcolor{\ourcolor}
            & \textbf{\methnam} & & \underline{88.6} & \textbf{60.7} &\textbf{89.2} & \textbf{71.1} & \textbf{65.1} & \textbf{24.7} & \textbf{88.4} & \underline{45.1} & \textbf{55.7} & \underline{68.1} & \textbf{65.7}\\
            \midrule 
            \multirow{6}{*}{\rotatebox{90}{\textbf{Average}}} &Continual-FT    & 42.1 & 70.5 & \textbf{92.2} & 80.1 & 54.5 & 59.1 & 19.8 & 78.3 & 41.0 & 38.1 & 42.3 & 56.2 \\
            & LwF \cite{li2017lwf} & 49.0 & 77.0 & \underline{92.1} & 85.9 & 66.5 & 67.2 & 20.9 & 84.7 & 44.6 & 45.5 & 50.5 & 62.2  \\
            & WiSE-FT \cite{wortsman2022robust} & 52.6 & 79.3 & 91.9 & 83.9  & 63.4 & 65.2 & 23.3 & 83.7  & 45.4 & 40.0 & 48.2 & 61.5\\
            & ZSCL \cite{zheng2023preventing} & 81.7& \textbf{91.3} & 91.1 & 91.0 & 82.9 & \underline{72.5} & \underline{33.6} & \underline{89.7} & \underline{53.3} & \textbf{62.8} & \textbf{69.9} &   74.5 \\
            & MoE Ad. \cite{yu2024boosting} & \underline{84.9} & 89.9 & 89.3 & \underline{91.4} & \underline{86.2} & 72.2 & 33.4 & 89.4 & \underline{53.3} & 61.4 & \textbf{69.9} & \underline{74.7} \\
            \rowcolor{\ourcolor}
            & \textbf{\methnam} & \textbf{86.0} & \underline{90.9} & 89.5 & \textbf{92.3} & \textbf{87.2} & \textbf{73.0} & \textbf{37.0} & \textbf{91.5} & \textbf{53.5} & \underline{62.4} & \underline{69.1} & \textbf{75.7}\\
            \midrule
            \multirow{6}{*}{\rotatebox{90}{\textbf{Last}}} &Continual-FT & 24.0& 67.3 &99.1 & 87.4 & 44.3 & 67.0 & 29.5 & 92.3 & 61.3 & 81.0 & \textbf{88.1} & 67.4 \\
            & LwF \cite{li2017lwf} & 34.6 & 69.6 & \underline{99.3} & 88.7 & 61.1 & 72.5 & 32.5 & 88.1 & 65.6 & 90.9 & \underline{87.9} & 71.9 \\
            & WiSE-FT \cite{wortsman2022robust} & 35.6 & 76.9 & \textbf{99.5} & 89.1 & 62.1 & 71.8 & 27.8 & 90.8 & 67.0 & 85.6 & 87.6 &  72.2 \\
            & ZSCL \cite{zheng2023preventing} & 78.2 & \textbf{91.1} & 97.6 & \underline{92.5} & 87.4 & \underline{78.2} & 45.0 & 92.3 & 72.7 & \textbf{96.2} & 86.3 &  83.4\\
            & MoE Ad. \cite{yu2024boosting}& \underline{84.1} & 88.5 & 94.0 & 91.8 & \underline{94.1} & 77.8 & \underline{50.4} & \underline{93.3} & \textbf{77.1} & 87.7 & 86.6 & \underline{84.1}\\
            \rowcolor{\ourcolor}
            & \textbf{\methnam} & \textbf{86.2} & \underline{91.0} & 96.8 & \textbf{93.6} & \textbf{96.2} & \textbf{79.6} & \textbf{52.7} & \textbf{96.7} & \underline{76.2} & \underline{93.6} & 78.7 & \textbf{85.6}\\
        \bottomrule
    \end{tabular}}
        \caption{Comparison with state-of-the-art methods on the MTIL benchmark with Order II. All methods use CLIP ViT-B/16. We highlight the best and second best with \textbf{bold} and \underline{underline} styles respectively.}
    \label{tab:mtil_full_2}
\end{table*}

\tit{Extended MTIL results.} To provide a detailed comparison of our approach with state-of-the-art methods on the MTIL benchmark, we present fine-grained results for each dataset in~\cref{tab:mtil_full_1}. Specifically, we report $\text{TR}^i$ (Transfer), $A^i$ (Average), and $A_T^i$ (Last). Additionally, \cref{tab:mtil_full_2} shows the corresponding results for the MTIL benchmark under Order-II~\cite{zheng2023preventing}. Results from other methods are sourced from~\cite{yu2024boosting}.
\begin{table}[t]
\rowcolors{2}{lightgray}{}
\begin{center}
\begin{tabular}{lccccccc}
\toprule
\textbf{Model} & \textbf{\shortsplitimagenet} & \textbf{\shortsplitcars} & \textbf{\shortsplitcub} & \textbf{\shortspliteurosat} & \textbf{\shortsplitisic} & \textbf{Avg.} \\
\midrule
CLIP~\cite{radford2021learning} & \result{74.3} & \result{64.8} & \result{53.3} & \result{40.2} & \result{27.1} & \result{51.9} \\
AttriCLIP~\cite{wang2023attriclip} & \result{76.4} & \result{62.6} & \result{48.3} & \result{41.7} & \result{26.6} & \result{51.1} \\
MoE Ad.~\cite{yu2024moe} & \result{79.1} & \result{63.6} & \result{52.1} & \result{39.5} & \result{27.8} & \result{52.4} \\
ZSCL~\cite{zheng2023preventing} & \result{75.6} & \result{60.9} & \result{53.7} & \result{55.8} & \result{31.6} & \result{55.5} \\
CGIL~\cite{frascaroli2024CLIP} & \result{70.1} & \result{69.3} & \result{27.2} & \result{59.1} & \result{38.8} & \result{52.9} \\
\midrule
\textbf{\methnam\scriptsize{{\textcolor{gray}{(LoRA)}}}} & \resultb{82.4} & \resultb{78.0} & \resultb{66.1} & \result{66.6} & \result{46.3} & \result{67.9} \\
\textbf{\methnam\scriptsize{{\textcolor{gray}{(VeRA)}}}} & \result{81.6} & \result{77.8} & \result{66.0} & \resultb{67.6} & \resultb{49.4} & \resultb{68.5} \\
\bottomrule
\end{tabular}
\end{center}
\caption{The \ciltransfer on the tested benchmarks of zero-shot capable models with the ViT-B/16 as the backbone.}
\label{tab:vitb_transfer}
\end{table}

\begin{table}[t]
  \centering
\rowcolors{2}{lightgray}{}
\begin{center}
\begin{tabular}{lcccccc}
\toprule
\textbf{Model} & \textbf{\shortsplitimagenet} & \textbf{\shortsplitcars} & \textbf{\shortsplitcub} & \textbf{\shortspliteurosat} & \textbf{\shortsplitisic} & \textbf{Avg.}\\
\midrule
CLIP~\cite{radford2021learning} & \faa{74.2} & \faa{63.6} & \faa{53.3} & \faa{41.7} & \faa{28.6} & \result{52.3}\\
\midrule
CODA-P.~\cite{smith2023coda} & \result{75.4} & \result{32.0} & \result{67.3} & \result{63.1} & \result{44.9} &\result{56.5} \\
SLCA~\cite{zhang2023slca} & \result{77.0} & \result{67.7} & \resultb{84.7} & \result{88.7} & \result{59.2} & \result{75.5} \\
MoE Ad.~\cite{yu2024moe} & \resultb{82.0} & \result{65.7} & \result{55.8} & \result{72.0} & \result{37.1} & \result{62.5} \\
ZSCL~\cite{zheng2023preventing} & \result{80.6} & \result{66.8} & \result{53.4} & \result{74.6} & \result{37.8} & \result{62.7} \\
CGIL~\cite{frascaroli2024CLIP} & \result{80.8} & \result{83.3} & \result{75.9} & \result{94.4} & \result{71.4} & \result{81.1} \\
\midrule
\textbf{\methnam\scriptsize{{\textcolor{gray}{(LoRA)}}}} & \result{79.8} & \resultb{84.5} & \result{76.3} & \result{94.4} & \resultb{73.6} & \resultb{81.7} \\
\textbf{\methnam\scriptsize{{\textcolor{gray}{(VeRA)}}}} & \result{78.7} & \result{84.1} & \result{75.2} & \resultb{94.5} & \result{73.0} & \result{81.1} \\
\bottomrule
\end{tabular}
\end{center}
  \caption{The Final Avg. Accuracy on the tested benchmarks of models with the ViT-B/16 as the backbone.}
  \label{tab:vitb_results}
\end{table}

\tit{Class-IL with ViT-B/16.} Here, we provide additional results for the Class-Incremental Learning benchmark using the ViT-B/16 backbone. In~\cref{tab:vitb_results}, we provide results in term of Final Average Accuracy (FAA), while in \cref{tab:vitb_transfer}, we report the CI-Transfer metric. The results are consistent with those obtained using the ViT-L/14 backbone, with \methnam being the best-performing method overall.

\clearpage

\bibliography{bib}